# Comparing Deep Reinforcement Learning and Evolutionary Methods in Continuous Control

**Shangtong Zhang**[1], **Osmar R. Zaiane**[2]

[1,2] Dept. of Computing Science, University of Alberta
shangtong.zhang@ualberta.ca, osmar.zaiane@ualberta.ca

## Abstract

Reinforcement Learning and the Evolutionary Strategy are two major approaches in addressing complicated control problems. Both are strong contenders and have their own devotee communities. Both groups have been very active in developing new advances in their own domain and devising, in recent years, leading-edge techniques to address complex continuous control tasks. Here, in the context of Deep Reinforcement Learning, we formulate a parallelized version of the Proximal Policy Optimization method and a Deep Deterministic Policy Gradient method. Moreover, we conduct a thorough comparison between the state-of-the-art techniques in both camps fro continuous control; evolutionary methods and Deep Reinforcement Learning methods. The results show there is no consistent winner.

## 1 Introduction

The biological basis of reinforcement learning (RL) is the behavioral learning process of animals, where an individual learns knowledge by *trial-and-error*. However, the evolutionary strategy (ES) comes from the evolution of the species, where randomness happens at every time and only individuals with positive mutations can survive. Both mechanisms widely exist in nature and both are crucial to the development of the species. Similarly, both RL and ES have gained huge success in solving difficult tasks. However, different from nature, where individual learning and specie evolution are combined in a reasonable way, we have not had a unified framework to combine RL and ES until now. As a result, we need to understand the advantages and the weaknesses of both methods in order to select the proper approach when faced with different tasks. Our main contribution is a systematic comparison between the state-of-the-art techniques in both domains in continuous control problems. Deep learning has recently achieved great success in various domains, so in our comparison, we always use neural networks as function approximators, i.e. deep RL. Moreover, all the algorithms are implemented in a parallelized manner across multiple processes to exploit the advance of modern computation resources. Our other contribution is that we formulate the parallelized version of the Proximal Policy Optimization [Schulman *et al.*, 2017] and Deep Deterministic Policy Gradient [Lillicrap *et al.*, 2015] and showcase their power in continuous control tasks.

We conducted our comparison in terms of both wall time and environment time steps, i.e. the amount of interactions with the environment, to gain a better empirical understanding of the running speed and data efficiency of all the algorithms.

## 2 Compared Algorithms

Considering a *Markov Decision Process* with a state set $\mathcal{S}$, an action set $\mathcal{A}$, a reward function $r : \mathcal{S} \times \mathcal{A} \to \mathbb{R}$ and a transition function $p : \mathcal{S} \times \mathcal{A} \to (\mathcal{S} \to [0, 1])$, the goal of the agent is to learn an optimal policy to maximize the expected discounted return $G_t = \sum_{k=0}^{T-1} \gamma^k R_{t+k+1}$ at every time step, where $T$ denotes the terminal time and $\gamma$ is the discount factor. In the continuous control domain, an action

$a$ is usually a vector, i.e. $a \in \mathbb{R}^d$, where $d$ is the dimension of the action. We use $\pi$ and $\mu$ to represent a stochastic and a deterministic policy respectively, and we assume the policy is parameterized by $\theta$ in the following sections, which is a neural network.

## 2.1 Evolutionary Methods

Evolutionary methods solve the control problem by evolving the control policy. Different evolutionary methods adopt different mechanisms to generate *individuals* in a *generation*, where individuals with better performance are selected to produce the next generation.

**Covariance Matrix Adaptation Evolution Strategy (CMAES)**

In CMAES [Hansen and Ostermeier, 1996], the structure of the neural network is predefined, $\theta$ here only represents the weights of the network. Each generation $\mathcal{G}$ consists of many candidate parameters, i.e. $\mathcal{G} = \{\theta_1, \ldots, \theta_n\}$. Each $\theta_i$ is sampled from a multivariate Gaussian distribution in the following fashion: $\theta_i \sim \boldsymbol{\mu} + \sigma \mathcal{N}(\mathbf{0}, \boldsymbol{\Sigma})$, where $\mathcal{N}(\mathbf{0}, \boldsymbol{\Sigma})$ is a multivariate Gaussian distribution with zero mean and covariance matrix $\boldsymbol{\Sigma}$, $\boldsymbol{\mu}$ is the mean value of the search distribution and $\sigma$ is the overall standard deviation (aka step size). All the candidates in $\mathcal{G}$ are evaluated by the environment. According to the performance, a certain selection mechanism will select some candidates to update the sample distribution, i.e. update $\boldsymbol{\mu}$, $\sigma$ and $\boldsymbol{\Sigma}$, and the next generation is sampled from the updated distribution.

**NeuroEvolution of Augmenting Topologies (NEAT)**

NEAT [Stanley and Miikkulainen, 2002] has shown great success in many difficult control tasks. The basic idea of NEAT is to evolve both the structure and the weights of the network. Thus $\theta$ now represents both the weights and the structure of the network. At every generation, NEAT selects several best individuals to crossover. And their descendants, with various mutations, will form the next generation. NEAT introduced a *genetic encoding* to represent the network efficiently and used the *historical markings* to perform a crossover among networks in a reasonable way, which helped avoid expensive topological analysis. Furthermore, NEAT incorporates an *innovation number* to protect innovations, which allowed the network to evolve from the simplest structure.

**Natural Evolution Strategy (NES)**

In NES [Wierstra *et al.*, 2008], the structure of the network is given, $\theta$ here only represents the weights. NES assumes the population of $\theta$ is drawn from a distribution $p_\phi(\theta)$, parameterized by $\phi$, and aims to maximize the expected fitness $J(\phi) = \mathbb{E}_{\theta \sim p_\phi} F(\theta)$, where $F$ is the fitness function to evaluate an individual $\theta$. Salimans *et al.* [2017] instantiated the population distribution $p_\phi$ as a multivariate Gaussian distribution, with mean $\phi$ and a fixed covariance $\sigma^2 I$. Thus $J$ can be rewritten in terms of $\theta$ directly, i.e. $J(\theta) = \mathbb{E}_{\epsilon \sim \mathcal{N}(0,I)} F(\theta + \sigma \epsilon)$. In a fashion similar to policy gradient, we have $\nabla_\theta J(\theta) = \frac{1}{\sigma} \mathbb{E}_{\epsilon \sim \mathcal{N}(0,I)} F(\theta + \sigma \epsilon) \epsilon$. In practice, at every generation, the population $\{\epsilon_1, \ldots, \epsilon_n\}$ is drawn from $\mathcal{N}(0, I)$. The update rule for $\theta$ is, therefore, $\theta \leftarrow \theta + \alpha \frac{1}{n\sigma} \sum_{i=1}^n F(\theta + \sigma \epsilon_i) \epsilon_i$, where $\alpha$ is the step size and $n$ is the population size.

## 2.2 Deep Reinforcement Learning Methods

In deep RL methods, the structure of the network is predefined, $\theta$ simply represents the numeric weights of the neural network.

**Continuous Asynchronous Advantage Actor-Critic (CA3C)**

The goal of the actor-critic method is to maximize the value of the initial state $v_\pi(s_0)$, where $v_\pi(s)$ is the value of the state $s$ under policy $\pi$, i.e. $v_\pi(s) \doteq \mathbb{E}_\pi[\sum_{k=0}^{T-1} \gamma^k R_{t+k+1} \mid S_t = s]$. Let $q_\pi(s, \boldsymbol{a}) \doteq \mathbb{E}_\pi[\sum_{k=0}^{T-1} \gamma^k R_{t+k+1} \mid S_t = s, A_t = \boldsymbol{a}]$ denote the value of the state action pair $(s, \boldsymbol{a})$, we have $v_\pi(s) = \sum_{\boldsymbol{a} \in \mathcal{A}} \pi(\boldsymbol{a}|s) q_\pi(s, \boldsymbol{a})$. Our objective, therefore, is to maximize $J(\theta) = v_\pi(s_0)$. According to the policy gradient theorem [Sutton *et al.*, 2000], we have $\nabla_\theta J(\theta) = \mathbb{E}_\pi[\nabla_\theta log \pi(A_t|S_t, \theta)(R_{t+1} + \gamma v_\pi(S_{t+1}) - v_\pi(S_t))]$. The value function $v_\pi$ is updated in a semi-gradient TD fashion [Sutton and Barto, 1998]. In continuous action domain, the policy $\pi$ is often parameterized in the form of the probability density function of the multivariate Gaussian distribution, i.e. $\pi(\boldsymbol{a}|s, \theta) \doteq \mathcal{N}(\boldsymbol{\mu}(s, \theta), \boldsymbol{\Sigma}(s, \theta))$. In practice, setting the covariance matrix $\boldsymbol{\Sigma}(s, \theta) \equiv \boldsymbol{I}$ is a good choice to increase stability and simplify the parameterization. Beyond the standard advantage actor-critic method, Mnih *et al.* [2016] introduced asynchronous workers to gain uncorrelated data, speed up learning and reduce variance, where multiple workers were distributed across processes and every worker interacted with the environment separately to collect data. The computation of the gradient was also in a non-centered manner.

**Parallelized Proximal Policy Optimization (P3O)**

Schulman *et al.* [2015] introduced an iterative procedure to monotonically improve policies named Trust Region Policy Optimization, which aims to maximize an objective function $L(\theta)$ within a trust region, i.e.

$$\underset{\theta}{\text{maximize}} \quad L(\theta) = \mathbb{E}[\frac{\pi_\theta(\boldsymbol{a}_t|s_t)}{\pi_{\theta_{old}}(\boldsymbol{a}_t|s_t)} A_t]$$
$$\text{subject to} \quad \mathbb{E}[D_{KL}(\pi_\theta(\cdot|s_t), \pi_{\theta_{old}}(\cdot|s_t))] \leq \delta$$

where $A_t$ is the advantage function, $D_{KL}$ is the KL-divergence and $\delta$ is some threshold. In practice, solving the corresponding unconstrained optimization problem with a penalty term is more efficient, i.e. we maximize $L^{KL}(\theta) = L(\theta) - \beta \mathbb{E}[D_{KL}(\pi_\theta(\cdot|s_t), \pi_{\theta_{old}}(\cdot|s_t))]$ for some coefficient $\beta$. Furthermore, Schulman *et al.* [2017] proposed the clipped objective function, resulting in the Proximal Policy Optimization (PPO) algorithm. With $r_t(\theta) \doteq \frac{\pi_\theta(\boldsymbol{a}_t|s_t)}{\pi_{\theta_{old}}(\boldsymbol{a}_t|s_t)}$, the objective function of PPO is $L^{CLIP}(\theta) = \mathbb{E}[\min(r_t(\theta)A_t, clip(r_t(\theta), 1-\epsilon, 1+\epsilon)A_t)]$ where $\epsilon$ is a hyperparameter, e.g. $\epsilon = 0.2$. In practice, Schulman *et al.* [2017] designed a truncated generalized advantage function, i.e. $A_t = \delta_t + (\gamma\lambda)\delta_{t+1} + \cdots + (\gamma\lambda)^{T-t+1}\delta_{T-1}$ where $T$ is the length of the trajectory, $\delta_t$ is the TD error, i.e. $\delta_t = r_t + v_\pi(s_{t+1}) - v_\pi(s_t)$, and $\lambda$ is a hyperparameter, e.g. $\lambda = 0.97$. The value function is updated in a semi-gradient TD manner.

We parallelize the PPO mainly following the training protocol of A3C and the Distributed PPO (DPPO) method [Heess *et al.*, 2017]. In P3O, we distribute multiple workers across processes like A3C. Following DPPO, each worker in P3O also has its own experience replay buffer to store its transitions. The optimization that happens in the worker is solely based on this replay buffer. Different from DPPO, where the objective function was $L^{KL}(\theta)$, we use $L^{CLIP}(\theta)$ in P3O, as it was reported to be able to outperform $L^{KL}(\theta)$ [Schulman *et al.*, 2017]. Heess *et al.* [2017] reported that synchronized gradient update can actually outperform asynchronous update in DPPO. So in P3O we also used the synchronized gradient update. However our synchronization protocol is quite simple. We use A3C-style update with an extra shared lock for synchronization and all the gradients are never dropped, while DPPO adopted a quite complex protocol and some gradients may get dropped. Moreover, a worker in P3O only does one batch update based on all the transitions in the replay buffer as we find this can increase the stability. In DPPO, however, each worker often did multiple mini-batch updates.

**Distributed Deep Deterministic Policy Gradient (D3PG)**

Similar to CA3C, the goal of DDPG is also to maximize $J(\theta) = v_\mu(s_0)$. According to the deterministic policy gradient theorem [Silver *et al.*, 2014], we have $\nabla_\theta J(\theta) = \mathbb{E}_{\mu'}[\nabla_a q_\mu(s, \boldsymbol{a})|_{s=s_t, \boldsymbol{a}=\mu(s_t|\theta)} \nabla_\theta \mu(s|\theta)|_{s=s_t}]$, where $\mu'$ is the behavior policy. The behavior policy usually combines the target policy with some noise, i.e. $\mu'(s) = \mu(s) + \mathcal{N}$, where $\mathcal{N}$ is some random noise. Following [Lillicrap *et al.*, 2015], we use an Ornstein-Uhlenbeck process [Uhlenbeck and Ornstein, 1930] as the noise. To stabilize the learning process, Lillicrap *et al.* [2015] introduced experience replay and target network, resulting in the DDPG algorithm.

We formulate the distributed version of DDPG in analogy to P3O, except the experience replay buffer is shared among different workers. Each worker interacts with the environment, and acquired transitions are added to a shared replay buffer. At every time step, a worker will sample a batch of transitions from the shared replay buffer and compute the gradients following the DDPG algorithm. The update is synchronized and the target network is shared.

## 3 Experiments

### 3.1 Testbeds

We investigated three kinds of tasks for our evaluation. The first kind of task is the classical toy task to verify the implementation. The second kind of tasks needs careful exploration and the third kind of tasks involves rich dynamics. We used the *Pendulum* task as a representative of the classical toy task. Recently *MuJoCo* has become popular as a benchmark in the continuous control domain. However there have been many empirical results on *MuJoCo* in the community. In our comparison, we considered the *Box2D* [Catto, 2011] environment, especially the

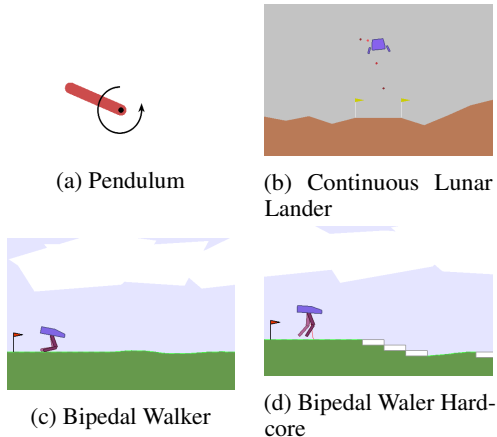

Figure 1: Four tasks. The background of the *Continuous Lunar Lander* was edited for better display.

*Continuous Lunar Lander* task, the *Bipedal Walker* task and the *Bipedal Waler Hardcore* task. *Continuous Lunar Lander* is a typical task that needs careful exploration. Negative rewards are continuously given during the landing so the algorithm can easily get trapped in a local minima, where it avoids negative rewards by doing nothing after certain steps until timeout. *Bipedal Walker* is a typical task that includes rich dynamics and *Bipedal Walker Hardcore* increases the difficulty for the agent to learn how to walk by putting some obstacles in the path. The three *Box2D* tasks have comparable complexity with *MuJoCo* tasks, and all the four tasks are free and available in OpenAI Gym. All the tasks have low-dimensional vector state space and continuous action. Some screenshots of the tasks are shown in **Figure 1**.

### 3.2 Performance Measurement

For evolutionary methods, we performed 10 test episodes for the best individual in each generation and averaged the test results to represent the performance at those time steps (wall time). For CA3C and P3O, an additional test process run deterministic test episodes continuously. We performed 10 independent runs for each algorithm and each task without any fixed random seed. Our training was based on episodes (generations), while the length (environment steps and wall time) of episodes is different. So the statistics are not aligned across different runs. We used linear interpolation to average statistics across different runs.

### 3.3 Policy Representation

Following [Schulman *et al.*, 2017], we used two two-hidden-layer neural networks to parameterize the policy function and the value function for all the evaluated methods except NEAT. To investigate the scalability of the algorithms in terms of the amount of the parameters, we tested small networks (hidden layers with 16 hidden units) and large networks (hidden layers with 64 hidden units) respectively. We always used the hyperbolic tangent activation function and the two networks did not have shared layers. The output units were linear to produce the mean of the Gaussian distribution (for P3O and CA3C) or the action (for D3PG, CMAES and NES). Moreover, we used the identity covariance matrix for P3O and CA3C, thus the entropy penalty was simply set to zero. For NEAT, the initial architecture had only one hidden unit, as NEAT allows the evolution to start from the simplest architecture.

### 3.4 Wall Time

We implemented a multi-process version of CMAES and NEAT based on *Lib CMA* [1] and *NEAT-Python* [2], while all other algorithms were implemented from scratch in *PyTorch* for a like-to-like comparison. For deep RL algorithms, the most straightforward parallelization method is to only parallelize data collection, while the computation of the gradients remains centralized. However this approach needs careful balance design between data collection and gradients computation. Moreover, it is unfriendly to data efficiency. So we adopted algorithm dependent parallelization rather than this uniform solution. We run all the algorithms in the same server [3]. Although P3O and D3PG can be accelerated by GPU, we did not introduce GPU in our experiments as most of the evaluated algorithms are not compatible with GPU. Although the relative wall time is highly dependent on the implementation, the comparison we did here can still give us some basic intuition. We made all the implementations publicly available [4] [5].

---

[1] https://github.com/CMA-ES/pycma
[2] https://github.com/CodeReclaimers/neat-python
[3] Intel® Xeon® CPU E5-2620 v4

### 3.5 Hyper-parameter Tuning

Due to the huge amount of the hyper-parameters and the complexity of the tasks, we can not afford a thorough grid search for all the tasks and hyper-parameters. We only did the grid search for one or two key hyper-parameters for each algorithms in the *Pendulum* task and tuned all other hyper-parameters empirically or used the default value of the package. For other tasks, we used same parameters as *Pendulum* without further tuning. This also gives us an intuition of the robusticity of the algorithms in terms of the hyper-parameters. For CA3C, D3PG and P3O, we used the Adam optimizer for both the policy function and the value function, and the initial learning rates for the policy function and the value function were the same. We searched the learning rate in $\{10^{-4}, \ldots, 10^{-1}\}$. We found $10^{-4}$ gave best performance for CA3C and D3PG while $10^{-3}$ gave best performance for P3O. For NES, we searched the variance $\sigma$ and the learning rate $\alpha$ in $\{10^{-2}, \ldots, 10^{0}\} \times \{10^{-3}, \ldots, 10^{0}\}$. Setting both $\sigma$ and $\alpha$ to 0.1 gave best performance. For CMAES, we searched the overall standard deviation $\sigma$ in $\{10^{-2}, \ldots, 10^{1}\}$ and found 1 gave the best performance.

### 3.6 Normalization

For all the evaluated methods, we normalized the state with running estimation of the mean and the variance. All the running statistics were shared and updated across parallelized workers. For CA3C, P3O and D3PG, we also normalized the reward in the same manner as the state. For NES and CMAES, we adopted the reward shaping as reported in [Salimans *et al.*, 2017].

### 3.7 Results

We reported the performance in terms of both the environment steps and wall time in **Figure 2** and **Figure 3**. All the curves were averaged over 10 independent runs. For each run the maximum environment step is set to $10^7$. As deep RL methods had larger variance than evolutionary methods, all the curves of CA3C, P3O and D3PG were smoothed by a sliding window of size 50.

---

[4]https://github.com/ShangtongZhang/DeepRL
[5]https://github.com/ShangtongZhang/DistributedES

## 4 Discussion

We can hardly say which algorithm is the best, as no algorithm can consistently outperform others across all the tasks in terms of all the measurements. However we can still learn some general properties of methods in different domains.

### 4.1 Final Performance

The relative final performance was quite task dependent. One interesting observation is that NEAT achieved a good performance level in the three *Box2D* tasks, but failed the simplest *Pendulum* task. Moreover, all the evolutionary methods solved the *Continuous Lunar Lander* task, but most deep RL methods appeared to get trapped in some local minima, which denotes that evolutionary methods are better at exploration than the deep RL methods. However when it comes to the two *Walker* tasks, where rich dynamics are involved, most deep RL methods worked better and many evolutionary methods appeared to get trapped in some local minima, which denotes that the deep RL methods can handle rich dynamics better than the evolutionary methods. In general, NEAT, D3PG and P3O are good choices.

### 4.2 Learning Speed

In the *Pendulum* task the comparison is clear, deep RL methods perform better in terms of both environment steps (data efficiency) and wall time than evolutionary methods. This conclusion still holds in the *Bipedal Walker* task, although there is an exception that D3PG runs quite slowly in this task. As many deep RL methods failed the *Continuous Lunar Lander* task and many evolutionary methods failed the *Bipedal Walker Hardcore* task, we did not take those two tasks into consideration in the comparison of the learning speed.

### 4.3 Stability

In all the experiments, deep RL methods appeared to have larger variance than evolutionary methods. This observation conforms with the fact that evolutionary methods perform better in tasks which need careful exploration. As evolutionary methods have better exploration, they are less sensitive to the initialization and randomness, resulting in better stability.

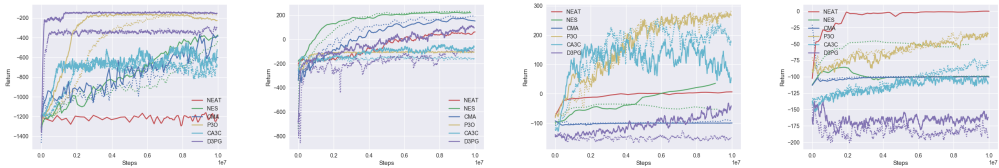

Figure 2: Performance in terms of environment steps. Left to right: *Pendulum, Continuous Lunar Lander, BipedalWalker, BipedalWalkerHardcore*. Solid line: large networks. Dashed line: small networks.

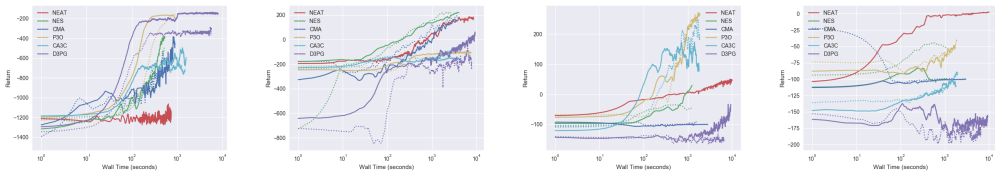

Figure 3: Performance in terms of wall time. Left to right: *Pendulum, Continuous Lunar Lander, BipedalWalker, BipedalWalkerHardcore*. Solid line: large networks. Dashed line: small networks. X-axis is in *log* scale

### 4.4 Scalability

In most experiments, the performance of deep RL methods improved with the increase of the network size. However sometimes small networks gave better performance for evolutionary methods than larger networks, e.g. NES with a small network reached a reasonable performance level in the *Bipedal Walker Hardcore* task, however it almost did not learn anything with a large network. In general our experiments showed that deep RL methods can make better use of complexity function parameterization and scale better to large network than evolutionary methods.

## 5 Future Work

Although our testbeds include several representative tasks, our comparison is still limited in tasks with low dimensional vector state space. With the popularity of the Atari games, image input has become a new norm in the discrete action domain. Our future work will involve continuous control tasks with image input.

## 6 Related Work

Taylor *et al.* [2006] compared NEAT with SARSA, but their work was limited on the football game Keepaway, which is a simple discrete control task. This comparison was extended later on by Whiteson *et al.* [2010], where new criterion was involved but the comparison was still limited to primitive RL methods where deep neural networks were not included. Duan *et al.* [2016] compared various deep RL methods with some evolutionary methods. However they did not include the NEAT paradigm and there has been exciting progress in both RL and ES, e.g. [Schulman *et al.*, 2017] and [Salimans *et al.*, 2017] after their work. Therefore these comparisons are not representative enough of the current state of the art. Moreover our comparison is conducted mainly on different tasks and measurements and focused on parallelized implementations.

### Acknowledgments

The authors thank G. Zacharias Holland and Yi Wan for their thoughtful comments.